%% file: arxiv.tex
\documentclass{msas}

\renewcommand{\today}{}

\usepackage[backend=bibtex,style=numeric,minnames=1,maxnames=5,maxcitenames=1,maxbibnames=99,natbib=true,doi=false]{biblatex}

\AtEveryBibitem{
\ifentrytype{inproceedings}{
 	\clearlist{address}
 	\clearlist{publisher}
 	\clearname{editor}
 	\clearlist{organization}
 	\clearfield{url}  
 	\clearfield{doi}  
 	\clearfield{pages}  
 	\clearlist{location}
 }{}
 \ifentrytype{article}{
	 \clearfield{url}
 }
 }
\usepackage[utf8]{inputenc} %
\usepackage[T1]{fontenc}    %
\usepackage{hyperref}       %
\usepackage{url}            %
\usepackage{booktabs}       %
\usepackage{amsfonts}       %
\usepackage{nicefrac}       %
\usepackage{microtype}      %
\usepackage{csquotes}
\usepackage{graphicx}
\usepackage{caption}
\usepackage{subcaption}
\usepackage{float}
\usepackage{xcolor}
\usepackage{cleveref}
\usepackage{tikz}
\usetikzlibrary{positioning}
\usetikzlibrary{shapes.arrows, fadings, shapes}
\usetikzlibrary{arrows}
\usepackage{algorithm}
\usepackage[noend]{algpseudocode}

\definecolor{nice-red}{HTML}{E41A1C}
\definecolor{nice-orange}{HTML}{FF7F00}
\definecolor{nice-yellow}{HTML}{FFC020}
\definecolor{nice-green}{HTML}{4DAF4A}
\definecolor{nice-blue}{HTML}{377EB8}
\definecolor{nice-purple}{HTML}{984EA3}
\definecolor{nice-grey}{HTML}{6C7A89}
\definecolor{nice-pink}{HTML}{DB5A6B}

\newcommand{\norm}[1]{\left\lVert#1\right\rVert}

\correspondingauthor{jenshe@alumni.stanford.edu, jayesh.gupta@microsoft.com}

\author[1]{Jennifer She}
\author[2]{Jayesh K. Gupta}
\author[1]{Mykel J. Kochenderfer}
\affil[1]{Stanford University}
\affil[2]{Microsoft}

\title{Agent-Time Attention for Sparse Rewards Multi-Agent Reinforcement Learning}

\begin{abstract}
  Sparse and delayed rewards pose a challenge to single agent reinforcement learning. This challenge is amplified in multi-agent reinforcement learning (MARL) where credit assignment of these rewards needs to happen not only across time, but also across agents. We propose Agent-Time Attention (ATA), a neural network model with auxiliary losses for redistributing sparse and delayed rewards in collaborative MARL. We provide a simple example that demonstrates how providing agents with their own local redistributed rewards and shared global redistributed rewards motivate different policies. We extend several MiniGrid environments, specifically MultiRoom and DoorKey, to the multi-agent sparse delayed rewards setting. We demonstrate that ATA outperforms various baselines on many instances of these environments. Source code of the experiments is available at \url{https://github.com/jshe/agent-time-attention}.

\end{abstract}

\begin{document}
\maketitle

\input{01_introduction.tex}

\input{02_background.tex}

\input{03_methodology.tex}

\input{04_experiments.tex}

\input{05_conclusion.tex}
\printbibliography

\end{document}

%% file: 01_introduction.tex
\section{Introduction}

Cooperative multi-agent reinforcement learning (MARL) where a team of agents learn coordinated policies optimizing global team rewards has been extensively studied in recent years \cite{oroojlooyjadid2019review,hernandez2019survey}, and find potential applications in a wide variety of domains like robot swarm control \cite{huttenrauch2019deep,arques2020swarm}, coordinating autonomous drivers \cite{palanisamy2020multi,zhou2020smarts}, network routing \cite{ye2015multi,bouton2021coordinated}, etc.
Although cooperative MARL problems can be framed as a \emph{centralized} single-agent, with the team as that actor with the joint action space, such an approach doesn't scale well.
Joint action space grows exponentially with number of agents in such scenarios. 
Moreover, due to real world constraints on communication and observability, such framing is often not useful for a large number of real world applications.
Unfortunately, simply independently learning \emph{decentralized} policies based on local observations result into unstable learning and convergence issues due to non-stationarity from simultaneous exploration \cite{gupta2017cooperative,terry2020revisiting}.
This has resulted in MARL methods focusing on the \emph{centralized training decentralized execution} (CTDE) paradigm, where during training decentralized polices can have access to extra state information during training but not during evaluation.

Sparse and delayed rewards are difficult for reinforcement learning (RL) because the number of possible trajectories grows exponentially with time horizon and this makes attributing rewards to intermediate observations and actions exponentially more difficult \cite{dann2015sample,jiang2018open}. 
One approach to improve learning is to supply additional rewards through reward shaping in order to transform sparse delayed rewards problems into dense ones \cite{ng1999policy}.
However, reward shaping is often difficult because it requires environment-specific knowledge. 
For single agent problems, frameworks like RUDDER and SECRET \cite{NEURIPS2019_16105fb9, ijcai2020-368} allow constructing neural network models for reward redistribution; learning how sparse delayed rewards can be transformed into dense rewards for effective policy optimization.

Unfortunately, in the multi-agent reinforcement learning (MARL) setting, sparse and delayed rewards have not been explored extensively \cite{hernandez2019survey}.
Existing cooperative MARL methods instead are focused on the problem of deducing an agent's contribution to the overall team's success, assuming access to dense global team rewards.
These methods can take various forms, be it \emph{implicit} like \cite{pmlr-v80-rashid18a,sunehag2018value,mahajan2019maven,son2019qtran,zhang2020multi} where the global state-action value is decomposed as aggregation of each agent's state-action value while assigning the shared global rewards to each agent based on their actions, or \emph{explicit} such as COMA~\cite{foerster2018counterfactual} and LIIR~\cite{NEURIPS2019_07a9d3fe} based on computing difference rewards \cite{wolpert2002optimal} against particular reward baselines.
Implicit methods often encounter limitations in expressiveness with no clear strategy for continuous action domains, while explicit methods face limitations on reasoning about individual effect of individual agent actions on the shared global rewards.

In this work we instead focus on multi-agent domains with delayed or sparse global rewards.
Therefore any MARL framework will require reasoning about both the contribution's from different agents as well as team's actions in the past i.e. solve the \emph{credit assignment} problem along both the agent and time axes. 
We focus on an improving explicit method to ensure applicability on continuous action problems. 
To this end, we propose Agent-Time Attention (ATA), a neural network model trained on auxiliary losses for redistributing global, sparse and delayed team rewards across both time and agents into dense, local agent rewards. 
This model can be applied on top of different single agent RL methods such as Q-learning and policy gradient methods without additional modifications under the CTDE paradigm. 
We perform a pedagogical experiment on a multi-agent one-dimensional coin environment to emphasize the importance of holistically reasoning about credit assignment along both agent and time axes at the same time. 
In order to further evaluate ATA's performance, we extend several MiniGrid environments, specifically MultiRoom and DoorKey \cite{gym_minigrid}, to the multi-agent sparse rewards setting.
This enables us to test ATA on environments with long horizons, where sparse delayed rewards present the most difficulty.
We found that simply extending standard policy gradient methods with ATA, not just outperforms baselines that just do credit assignment on either agent or time axis, but also their straightforward combinations where they still reason about credit-assignment along these two axes separately.
Finally, we conduct extensive ablation studies to provide better insights into the various components of our approach.

%% file: 02_background.tex
\section{Background}

We consider an extension of the Markov Decision Process to the fully collaborative, partially observable  multi-agent setting, formalized as $\{n, S, A, \mathcal{T}, \mathcal{R}, \Omega, \mathcal{O}, \gamma\}$. $n$ is the number of agents, $S$ is the global state space, and $A$ and $\Omega$ are the action space and observation space of each agent. The transition function is $\mathcal{T}: S \times A^n \rightarrow S$, the reward function is $\mathcal{R}: S \times A^n \rightarrow \mathbb{R}$, the local observation function for each agent is $\mathcal{O}: S \times A \rightarrow \Omega$, and the discount factor is $\gamma \in [0, 1)$. Each agent has a stochastic policy $\pi_i(a_i | \tau_i)$ conditioned on its action-observation history $\tau_i \in (\Omega \times A)^*$. The agents' joint policies $
\mathbf{\pi}$ induces a value function $V^{\mathbf{\pi}}(s_t) = \mathbb{E}[R_t|s_t]$ and action-value function $Q^{\mathbf{\pi}}(s_t, \mathbf{a}_t) = \mathbb{E}[R_t|s_t, \mathbf{a}_t]$, where $R_t=\sum_{l=0}^\infty\gamma^l r_{t+l}$ is the discounted return. The advantage function is defined as $Q^{\mathbf{\pi}}(s_t, \mathbf{a}_t) - V^{\mathbf{\pi}}(s_t)$.

\subsection{Policy Gradient and Actor-Critic}
Policy gradient methods are one class of RL methods which are extensible to continuous action space problems not addressable by Q-learning methods. The classic policy gradient method for a single agent MDP is REINFORCE \cite{williams1992simple}, where a policy $\pi_{\theta}$ optimizes the objective 
\begin{align}
\mathbb{E}[\sum_{t=0}^T R_t \nabla_{\theta} \log \pi_{\theta} (a_t | s_t)].
\end{align}

A common variant replaces $R_t$ in the objective with an advantage function estimate $R_t-\hat{V}(s_t)$, for a state-value function estimate $\hat{V}$ in order to reduce variance.
Actor-Critic \cite{konda2000actor} replaces $R_t$ with  $\hat{Q}(s_t, a_t)$ for an action-value function estimate $\hat{Q}$, or an advantage function estimate $\hat{Q}(s_t, a_t) - \hat{V}(s_t)$. These estimates $\hat{V}$ and $\hat{Q}$ (the "critic") are learned along with original $\pi_\theta$ (the "actor").

The simplest extension of these methods to the multi-agent partially observable scenario is applying them independently to each agent \cite{gupta2017cooperative,foerster2018counterfactual} where actor $i$, or $\pi_{\theta i}(a_{ti}|\tau_{ti})$, optimizes the policy gradient objective defined by critic $i$ describing $\hat{V}_{i}(\tau_{ti})$ or $\hat{Q}_{i}(\tau_{ti}, a_{ti})$, which in turn learns to approximate $\mathbb{E}[R_t | \tau_{ti}]$ or $\mathbb{E}[R_t | \tau_{ti}, a_{ti}]$. 
Note that this doesn't resolve the credit-assignment problem between agents and the global team reward is used as the agent reward. 
COMA \cite{foerster2018counterfactual} modifies independent actor critic (IAC) by replacing the action-value function estimate with a centralized $\hat{Q}_{_\mathrm{COMA}}(s_t, \mathbf{a}_t)$ approximating $\mathbb{E}[R_t|s_t, \mathbf{a}_t]$ and the advantage function approximate for agent $i$ with $\hat{Q}_{_\mathrm{COMA}}(s_t, \mathbf{a}_t) - \sum_{a_i'} \pi_{\theta i}(a_i' |\tau_{ti}) \hat{Q}_{_\mathrm{COMA}}(s_t, (\mathbf{a}_{t\sim i}, a_i'))$, comparing the the expected return of all chosen agent actions to one that marginalizes out agent $i$'s action.
LIIR \cite{NEURIPS2019_07a9d3fe} builds on IAC by augmenting global rewards $r_t$ with learned local rewards defined by $ r_{\text{LIIR}}(\tau_{ti}, a_{ti})$, which is trained using a separate centralized critic $\hat{V}_{\textrm{LIIR}}(s_t, \mathbf{a}_t)$ in order to optimize the expected return under the original setting. In both methods, sparse rewards are not explicitly dealt with and as a result, their challenges carry over to training components like $\hat{Q}_{_\mathrm{COMA}}$ and $\hat{V}_{\textrm{LIIR}}$.

\subsection{Single-agent Reward Redistribution}
In the single agent setting, sparse and delayed rewards pose a significant problem, specifically in long horizon settings because the number of possible trajectories increases exponentially in horizon length $T$, and as a result, so does any Monte Carlo approximation of the expected return, enforcing the number of required samples to increase exponentially with $T$. 
Temporal-difference methods which approximate value functions using bootstrapping will be slow to learn due to the number of steps it takes for a delayed reward to propagate backwards to initial timesteps, and the exponential decay of the reward that occurs in the process.

Reward redistribution methods \cite{NEURIPS2019_16105fb9} have been one alternative to reward shaping for sparse and delayed reward problems. 
These methods replace sparse $r_t$ with dense redistributed rewards $\Tilde{r}_t$ from a learned model. This reward redistribution model is trained alongside the underlying RL algorithm and induces a new MDP for the algorithm. 
RUDDER \cite{NEURIPS2019_16105fb9} learns a reward redistribution model realized as an LSTM network \cite{10.1162/neco.1997.9.8.1735}, which learns to predict $R_T$, and optionally $R_t$ given $\{s_0, a_0, ..., s_t, a_t\}$. $\Tilde{r}_{t_{\mathrm{RUDDER}}}$ are defined using various contribution methods such as difference in consecutive return predictions $\hat{R}_{t+1} - \hat{R}_t$, as well as gradient methods like integrated-gradients. 
The paper shows that the induced MDP's are return-equivalent under perfect $\hat{R}_T$ assumption, and share the same optimal $\pi$ as the original MDP as a result.

\subsection{Transformer}
\label{subsec:transformer}
The transformer architecture \cite{vaswani2017polosukhin} has recently come to dominate the deep learning literature For sequential prediction tasks, 
Attractive elements of the architecture include its ability to be parallelized, its interpretability, and its ability to capture long term dependencies without being susceptible to vanishing gradients common in recurrent networks.

Key operation in the transformer is the attention mechanism.
Self-attention, parametrized by matrices $(W_k, W_q, W_v)$ of size $\mathbb{R}^{d_i \times d_k}$, is the most popular attention mechanism used to estimate the relation between sequence elements as non-linear similarity scores between all pairs of sequence elements.
Any restriction on the computational window is applied as a mask $M_c$ to the result of pairwise similarity computations.
More formally, given the input sequence in matrix form $X = (x_t)_{t=0,\ldots,T} \in \mathbb{R}^{T\times d_i}$ and the output sequence $Z = (z_t)_{t=0\ldots,T} \in \mathbb{R}^{T \times d_k}$ after application of self-attention, we have:
\begin{align}
    Z &= \text{softmax}\left(\frac{M_c \odot (QK^T) - C(1 - M_c)}{\sqrt{d_k}}\right) V \\
    Q &= XW_q\,\quad K = XW_k\,\quad V = XW_v\,\quad M_c \in \{0, 1\}^{T\times T}
\end{align}
where $\odot$ is the Hadamard product and $C$ is a large constant ($\approx 10^9$).

Another way to view this output representation is as a linear combination of the values of other elements: $z_t = \sum_{i=0}^T \alpha_{i} v_i$ where $\alpha_t$ represents the normalized attention weights for the predictions at timestep $t$ while summing to 1.

Attention mechanisms and transformers have come to adoption in MARL as well.
The focus is usually on learning some sort of cross agent features through attention such as extracting relevant information of each agent from other agents \cite{iqbal2019actor}, often combined with graph networks \cite{jiang2019graph,niu2021multi,li2021deep}. 
More recently, transformers have even seen application in framing the sequential decision making for reinforcement learning as a sequence learning problem~\cite{chen2021decision}.

Transformers have also seen success at the reward redistribution problem.
An alternative model to RUDDER was introduced as SECRET \cite{ijcai2020-368}. 
Rather than using LSTMs, SECRET's reward redistribution model is instead realized as a transformer decoder network \cite{vaswani2017polosukhin} which learns to predict the sign of $r_t$ at every timestep. 
Since the goal is to assign credit, the model is not allowed to peek into the future and the computational window of each element in the sequence is restricted to the past elements in the sequence.
The mask $M_c$ is therefore a lower triangular binary matrix and is often termed as a causal mask.
$\Tilde{r}_{t_{\mathrm{SECRET}}}$ is the sum of future $r_l$'s that $(s_t, a_t)$ contribute to, determined by the model's attention on timestep $t$ from $l$, $\alpha_{t\leftarrow l}$. 
While this method ensures return equivalence, $\alpha_{t\leftarrow l}$ enforces some restrictions on the possible redistributed rewards.

\citet{xiao2022agent} was another concurrent work that explored the idea of multi-agent reward redistribution for MARL.

%% file: 03_methodology.tex
\section{Methodology}

\begin{figure*}
    \centering
    \includegraphics[width=0.98\textwidth]{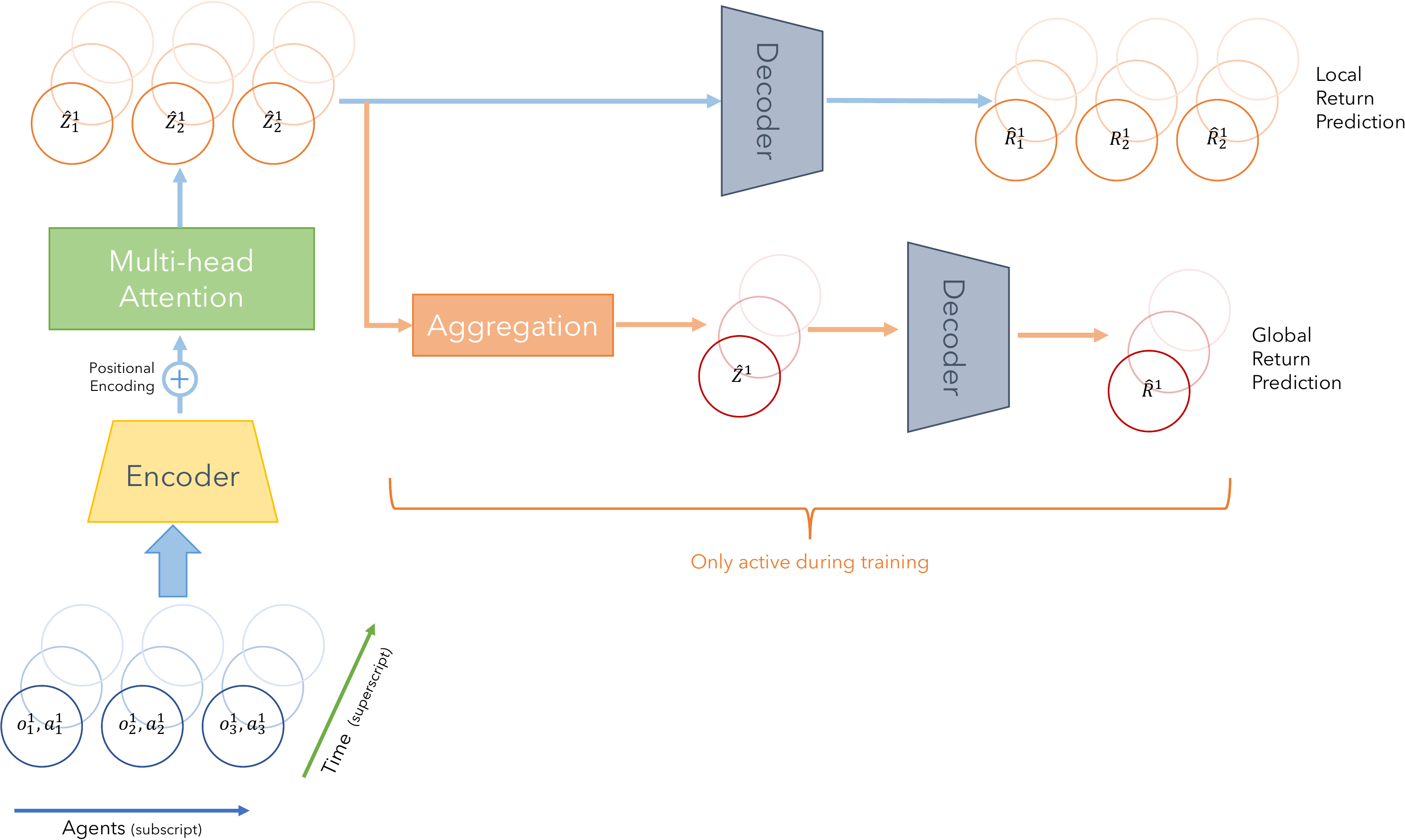}
    \caption{Network architecture of Agent-Time Attention (ATA) for Reward Redistribution. The \textcolor{nice-blue}{blue arrows} are active during both training and execution while \textcolor{nice-orange}{orange arrows} are active only during training. ATA first encodes individual agent observation-actions to a latent space. Together with positional encoding ATA applies a single layer of multi-head attention \cite{vaswani2017polosukhin} to another latent space. These are directly decoded to predict individual agent returns at particular time ($\hat{R}_i^t$ for $i$-th agent at $t$ timestep). We aggregate features in latent space across agents and decode them to predict global team returns at particular time ($\hat{R}^t$).}
    \label{fig:ata}
\end{figure*}

Our paper's hypothesis is that sparse rewards MARL requires reasoning about credit-assignment across both agents and time.
We are motivated by reward redistribution work in single-agent RL and wish to extend those ideas to the multi-agent case.

\subsection{MARL Reward Redistribution}

The original RUDDER ~\cite{NEURIPS2019_16105fb9} paper described reward redistribution as a two-step process. 
The first step is learning a function $g$ that predicts the expected return for a given observation-action sequence.
The second step is \emph{contribution analysis} that determines how much an observation-action pair contributes to the final prediction.

In the multi-agent case too, the basic principles behind reward redistribution remains the same.
However, there is an additional demand to redistribute rewards not just along time but also across agents.
Inspired by success of self-attention at \emph{implicit} credit assignment between agents in LICA \cite{zhou2020learning} and transformers at deep learning problems \cite{tay2020efficient,khan2021transformers,fournier2021practical} we propose a transformer based architecture for multi-agent return prediction which we term as Agent-Time Attention (ATA).
The network architecture is described in \Cref{fig:ata}.
The transformer details are as described in \Cref{subsec:transformer}. 
Observation-action pairs at each timestep $t$ for each agent $i$ result into the return predictions of $\hat{R}_i^t$ during execution.

The data for training this model is collected during policy optimization.
The reader might observe that in cooperative MARL we only have global team rewards from the environment.
To be able to train this architecture in a similar manner as RUDDER, we aggregate the features after the transformer layer, $\hat{Z}_1^t, \ldots \hat{Z}_n^t$, into $\hat{Z}^t$ before decoding with the same linear layer as used for local return predictions during execution.
The training loss is:
\begin{equation}
    \ell = (1 - \lambda) \norm{\hat{R}^t - R^t}^2 + \lambda \norm{\hat{R}_i^t - R_i^t}^2
    \label{eq:loss}
\end{equation}
where $R^t$ and $R_i^t$ are Monte Carlo returns estimated from the actual rollouts collected during policy optimization.

As recommended by RUDDER, we use \emph{differences in return predictions} for contribution analysis. The agent $i$'s reward at timestep $t$ is therefore:
\begin{equation}
    \tilde{r}_i^t = \hat{R}_i^{t+1} - \hat{R}_i^{t}
    \label{eq:diff}
\end{equation}
In our initial tests, we too found that this performed better than other contribution analysis methods like integrated gradients \cite{sundararajan2017axiomatic} or layer-wise relevance propagation \cite{bach2015pixel}.
For example, integrated gradients are more computationally expensive and result in many possible redistributed rewards when the reward redistribution model overfits.
The attention-weighed method from SECRET is also too constrained in that it enforces $\Tilde{r}_t = \alpha R_t$, where $\alpha \geq 0$. %
Our training algorithm is described in \Cref{algo:train}.

Our MARL reward redistribution model has the flexibility to provide agents with individual rewards in place of global rewards without requiring a MARL-specific policy optimization method. 
From past literature, global rewards tend to result in lazier agent behaviors and purely individual rewards tend to result in more selfish, potentially greedy agent behaviors. 
A balance can be achieved under different scenarios by tweaking the hyperparameter $\lambda$ in \Cref{eq:loss}.

\makeatletter
\def\BState{\State\hskip-\ALG@thistlm}
\makeatother

\begin{algorithm}
\caption{Reward Model Training and Reward Redistribution}\label{algo:train}
\begin{algorithmic}[1]
\Procedure{Train Model}{$g_\theta$} 
\State \Comment{$g_\theta$ is the reward redistribution model.}
\State $\textit{loss} \gets \infty$
\While {$\textit{loss} > \textit{min\_loss}$}
\State $(o, a, r)_{i=1\ldots n}^{t=1 \ldots T} \sim \textit{buffer}$. \Comment{Sample from data buffer.}
\State $\hat{R}_{i=1\ldots n}^{t=1 \ldots T}, \hat{R}^{t=1 \ldots T} \gets g_\theta((o, a)_{i=1\ldots n}^{t=1 \ldots T})$.
\State $\textit{loss} \gets \ell{(r_{i=1\ldots n}^{t=1 \ldots T}, \hat{R}_{i=1\ldots n}^{t=1 \ldots T}, \hat{R}^{t=1 \ldots T})}$. \Comment{\Cref{eq:loss}}
\State ${\theta} \gets \textit{update}(\textit{loss}, \theta)$.
\EndWhile
\EndProcedure
\Procedure{Reward Redistribution}{$g_\theta$}
\State $\hat{R}_{i=1\ldots n}^{t=1 \ldots T} \gets {g_\theta((o, a)_{i=1\ldots n}^{t=1 \ldots T})}$.
\State $\Tilde{r}_{i=1\ldots n}^{t=1 \ldots T} \gets \textit{difference}(\hat{R}_{i=1\ldots n}^{t=1 \ldots T})$. \Comment{\Cref{eq:diff}}
\EndProcedure
\end{algorithmic}
\end{algorithm}

%% file: 04_experiments.tex
\section{Experiments}

StarCraft II Multi-agent Challenge (SMAC) provides a collection of reinforcement learning benchmarks designed specifically for multi-agent environments \cite{samvelyan2019starcraft}.
While it allows converting some of the tasks into sparse reward problems by enabling the \texttt{sparse} flag, it is a computationally expensive domain.
Most existing algorithms compete on the non-sparse versions of the tasks and can sparse multi-agent problems in general can require a lot of samples from the environment to see any progress. 
Moreover most tasks in the benchmark are of relatively short horizons with no easy way to change that limit. Explicit reward redistribution methods tend to have larger effect with longer horizons.
We therefore focus on more pedagogical experiments to understand the problem of multi-agent credit assignment and effect of different solution approaches.
Moreover, unlike many CTDE approaches we do not assume we have access to the global state information during training i.e. MARL has to work with partial observations only.
All experiment curves were averaged over $6$ seeds and the fill color represents the standard deviation.

\subsection{1D Coin Environment}
\label{subsec:1d_exp}

To understand, why RUDDER like reward redistribution of global team rewards $\tilde{r}^t$ is not enough for MARL, and why we need to think about redistribution across both time and agents, we construct a simple one dimensional coin collection environment.

The environment consists of a line of length $13$ with two agents and one coin generated at random positions on the line.
Each agent has a field of view of length 5 with itself at the center, and possibly containing the other agent or the coin.
An agent therefore only has partial observations.
Agents can move left or right to get to the coin. 
If an agent gets to the coin, the global team reward increments by $p_1$, and if both agents get to the coin at the same time, the global reward increments by $p_2$.
Once either agents gets to the coin, the coin is randomly re-generated at a different position. 
We set the episode length to $200$. 
The global reward is only provided at the very end of each episode, making this environment a delayed rewards problem.

We compare our ATA model to a baseline RUDDER model that takes as input concatenation of the observations and actions of all agents into a single input $(o_{0t}, a_{0t}, o_{1t}, ... )$
and then predicts the global team reward redistribution only.
We use independent policy gradient (IPG) to train the agent policies.
\begin{figure}
     \centering
     \begin{subfigure}[b]{0.45\columnwidth}
         \centering
         \includegraphics[width=\textwidth]{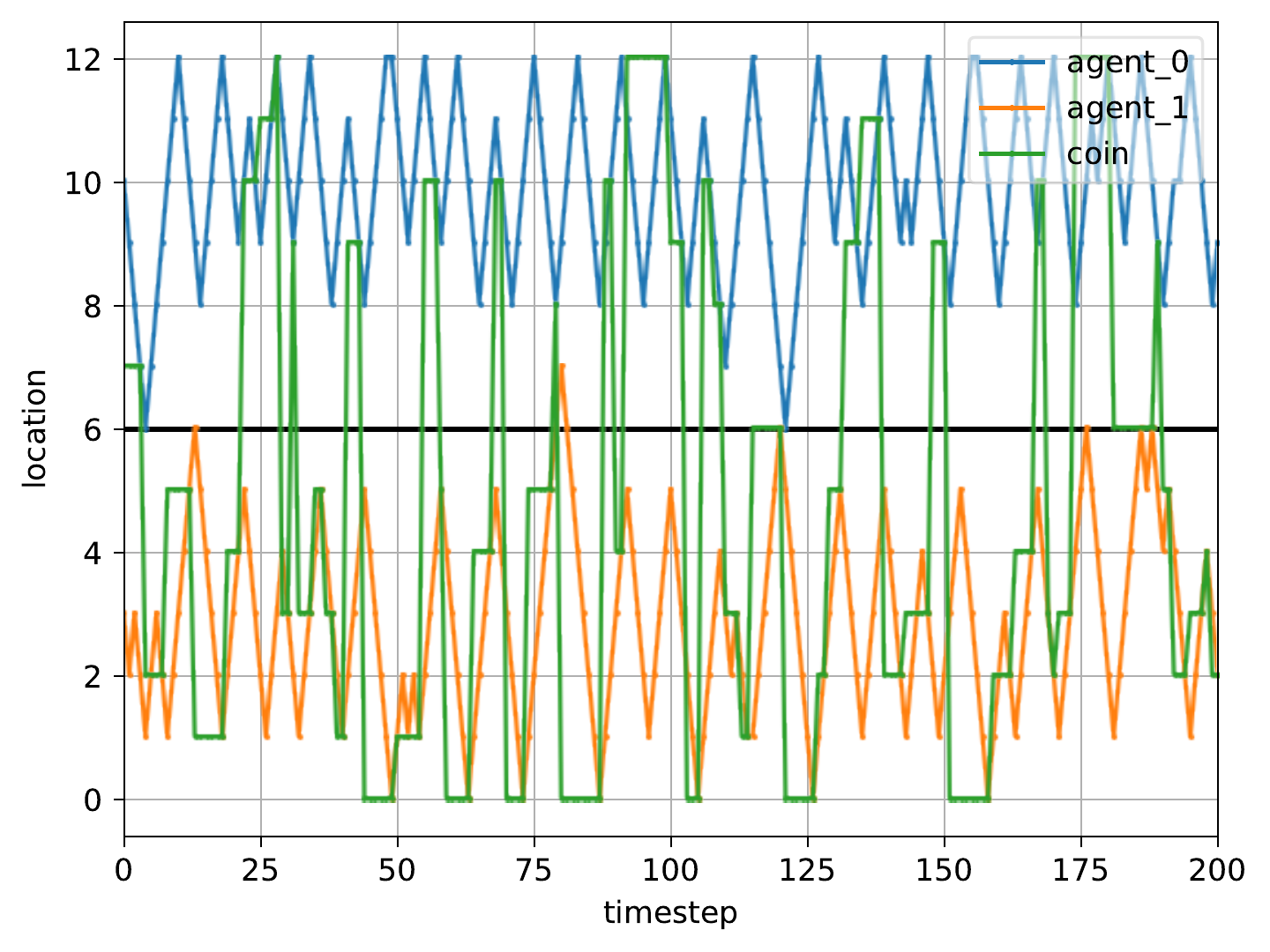}
         \caption{With reward redistribution only along time axis using RUDDER}
         \label{fig:1d_baseline}
     \end{subfigure}
     \begin{subfigure}[b]{0.45\columnwidth}
         \centering
         \includegraphics[width=\textwidth]{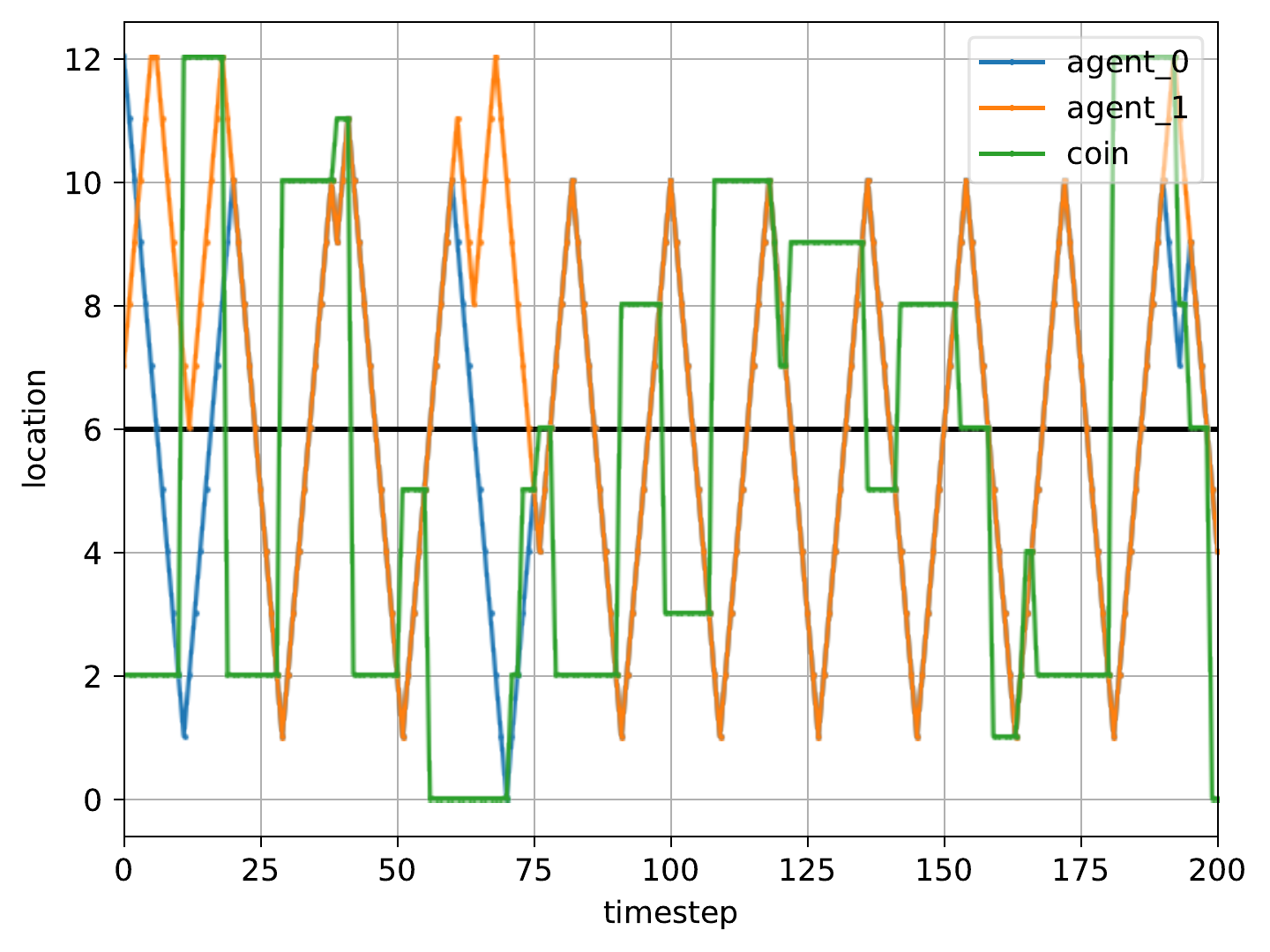}
         \caption{With reward redistribution along both agent and time axes using ATA}
         \label{fig:1d_test}
     \end{subfigure}
 \caption{1D Environment: sample episodes during training, of agents and coin positions over time. IPG-RUDDER-TR, IPG-ATA learn a RUDDER reward redistribution model and an ATA reward redistribution model respectively, each with a transformer decoder architecture. }
 \label{fig:1d_results}
\end{figure}

\begin{figure}
    \centering
    \includegraphics[width=0.65\columnwidth]{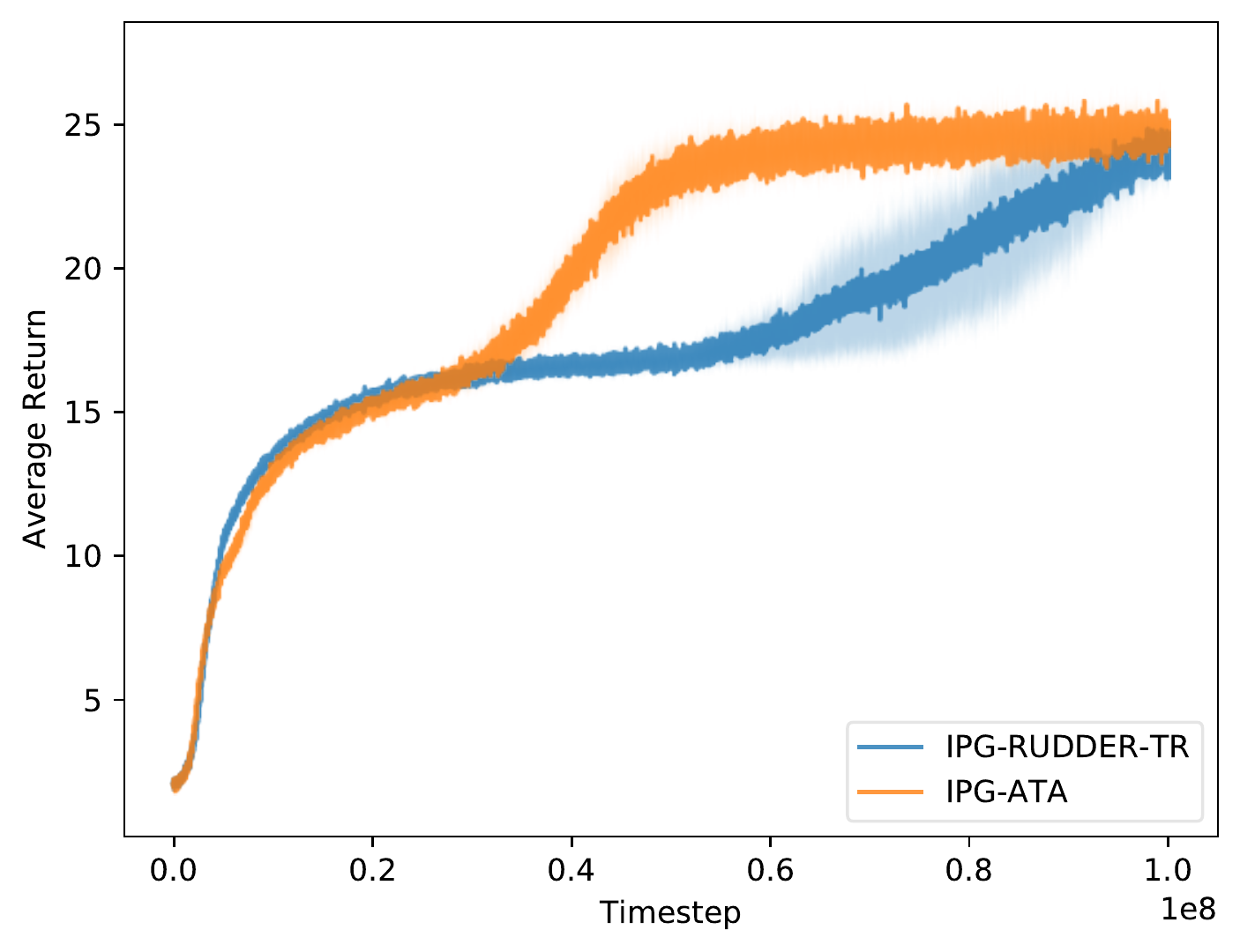}
    \caption{1D Environment: average return over training.}
    \label{fig:1d_results2}
\end{figure}

 \Cref{fig:1d_results} shows example episodes mid-training using these  reward redistribution methods for $p_1=0.25$ and $p_2=1$.
 Visually, we can see that the global team reward redistribution only along time encourages a lazy behavior of agents collecting coins in their own half of the space, while our reward redistribution model that does this redistribution along both agent and time axes can encourage the behavior where agents tend to move together, across the entire space. This is preferred specifically in the case where $p_2 > p_1$ because getting the coin together leads to higher returns.
 Although both methods converge to similar solutions in \Cref{fig:1d_results2}, ATA is much faster at getting there.

\subsection{MiniGrid Environments}
\label{subsec:2d_exp}

\begin{figure}
     \centering
     \begin{subfigure}[b]{0.48\columnwidth}
         \centering
         \includegraphics[width=0.95\columnwidth]{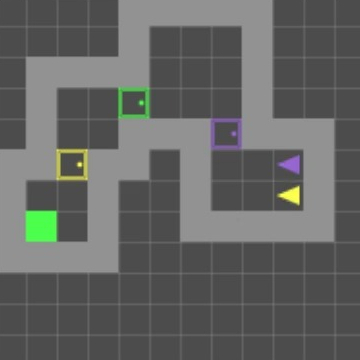}
         \caption{MultiRoom}
         \label{fig:2d_multiroom}
     \end{subfigure}
     \begin{subfigure}[b]{0.48\columnwidth}
         \centering
         \includegraphics[width=0.95\columnwidth]{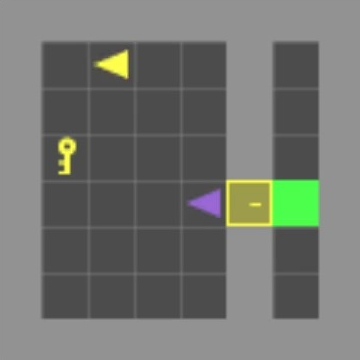}
         \caption{DoorKey}
         \label{fig:2d_doorkey}
     \end{subfigure}
 \caption{MiniGrid Environments: triangles are agents, green cells are coins and bordered cells are doors.}
 \label{fig:2d_envs}
\end{figure}

To test our models on a larger scale, we built sparse reward MARL environments based on MiniGrid \cite{gym_minigrid}. 
We specifically built off of the MultiRoom and DoorKey (Figure \ref{fig:2d_envs}) environments.
We reuse MiniGrid's default action space, which are: turn-left, turn-right, forward, pickup, toggle, drop, and done, of which only the first 4 are applicable to our specific environments.

\subsubsection{Environment Details}
The MultiRoom-inspired environment consists of a $12 \times 12$ grid containing $m$ rooms of size $2 \times 2$, $n$ agents and one coin (the goal item in the original MiniGrid) generated at random positions on the grid. Each agent has a field of view of $5 \times 5$ with itself at the center.
The policy needs to control the agents' movements, which are default MiniGrid actions, to get to the coin.

The DoorKey-inspired environment consists of a $8 \times 8$ grid containing a divider down the center with a locked door, $n$ agents, and one key generated randomly on one side of the divider, and one coin generated randomly on the other side of the divider. Again, each agent has  field of view of $5 \times 5$ with itself at the center, and we control the agents' movements to unlock the door and get to the coin.

In both environments, if $j$ agents get to the coin at the same time, the global reward increments by $p_j$. Once some agent gets to the coin, the coin is randomly re-generated at a different position. We again set the episode length to $200$. The global team reward is only provided at the very end of each episode, making these environments delayed rewards problems.

We experiment with $m=2, 3$ for MultiRoom and $n=2, 3$ across both MultiRoom and DoorKey. 
Empirically, we found that trends were generally consistent across different $p_j$, so we share results with $p_j = 1$ for all $j$. 
This means that optimal reward redistribution across agent and time will have to attribute the reward at the end of the rollout to the correct agents and the correct timestep.
We keep hyperparameters the same across these environments.

\subsubsection{Baselines}

We compare ATA trained along with independent policy gradient (IPG) to IPG without and reward redistribution, IPG with reward redistribution along time, a MARL method which does not explicitly deal with sparse and delayed rewards, as well as a combination of single-agent reward redistribution and MARL.

\begin{itemize}
\item \textbf{IPG \cite{williams1992simple}:} Each agent learns a separate policy network and value network, and the rewards for each agent are the global sparse rewards that come from the environment.
\item \textbf{COMA \cite{foerster2018counterfactual}:} Each agent learns a separate policy network but the value network is centralized and specific to MARL, for decomposing global values into agent-specific ones.
\item \textbf{IPG + RUDDER \cite{NEURIPS2019_16105fb9}:} A centralized RUDDER LSTM network is learned along with IPG, and learns to redistribute global rewards across timesteps given observations and actions of all agents.  Global redistributed rewards are used in place of original rewards in training IPG.
\item\textbf{ COMA \cite{foerster2018counterfactual} + RUDDER \cite{NEURIPS2019_16105fb9}:} A centralized RUDDER LSTM network is learned along with COMA, similar to the IPG + RUDDER setting.
\end{itemize}

\subsubsection{Discussion}
\begin{figure*}
     \centering
     \begin{subfigure}[b]{0.4\textwidth}
         \centering
         \includegraphics[width=\columnwidth]{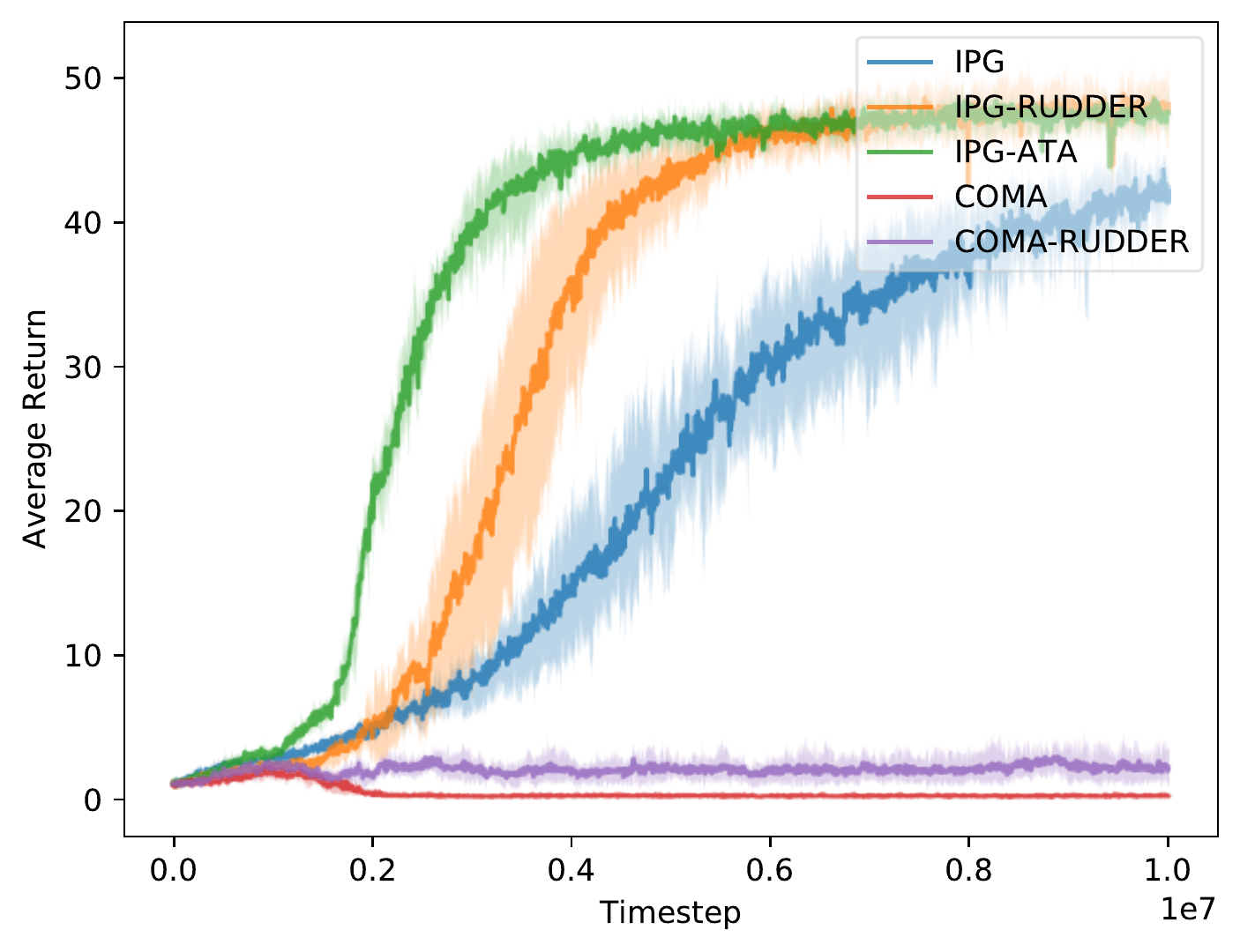}
         \caption{\#Agents = 2, \#Rooms = 2}
     \end{subfigure}
     \begin{subfigure}[b]{0.4\textwidth}
         \centering
         \includegraphics[width=\columnwidth]{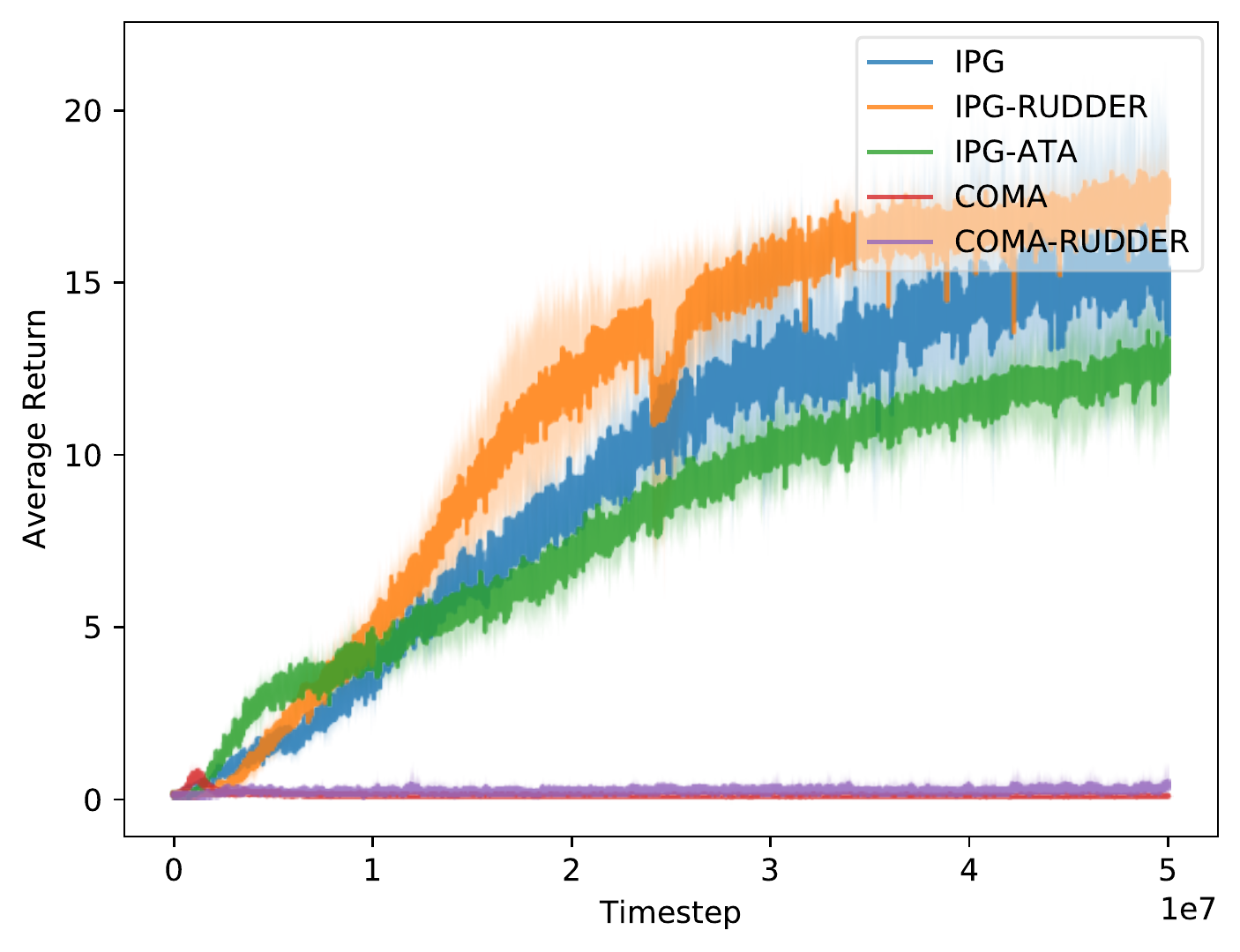}
         \caption{\#Agents = 2, \#Rooms = 3}
     \end{subfigure}
     \begin{subfigure}[b]{0.4\textwidth}
         \centering
         \includegraphics[width=\columnwidth]{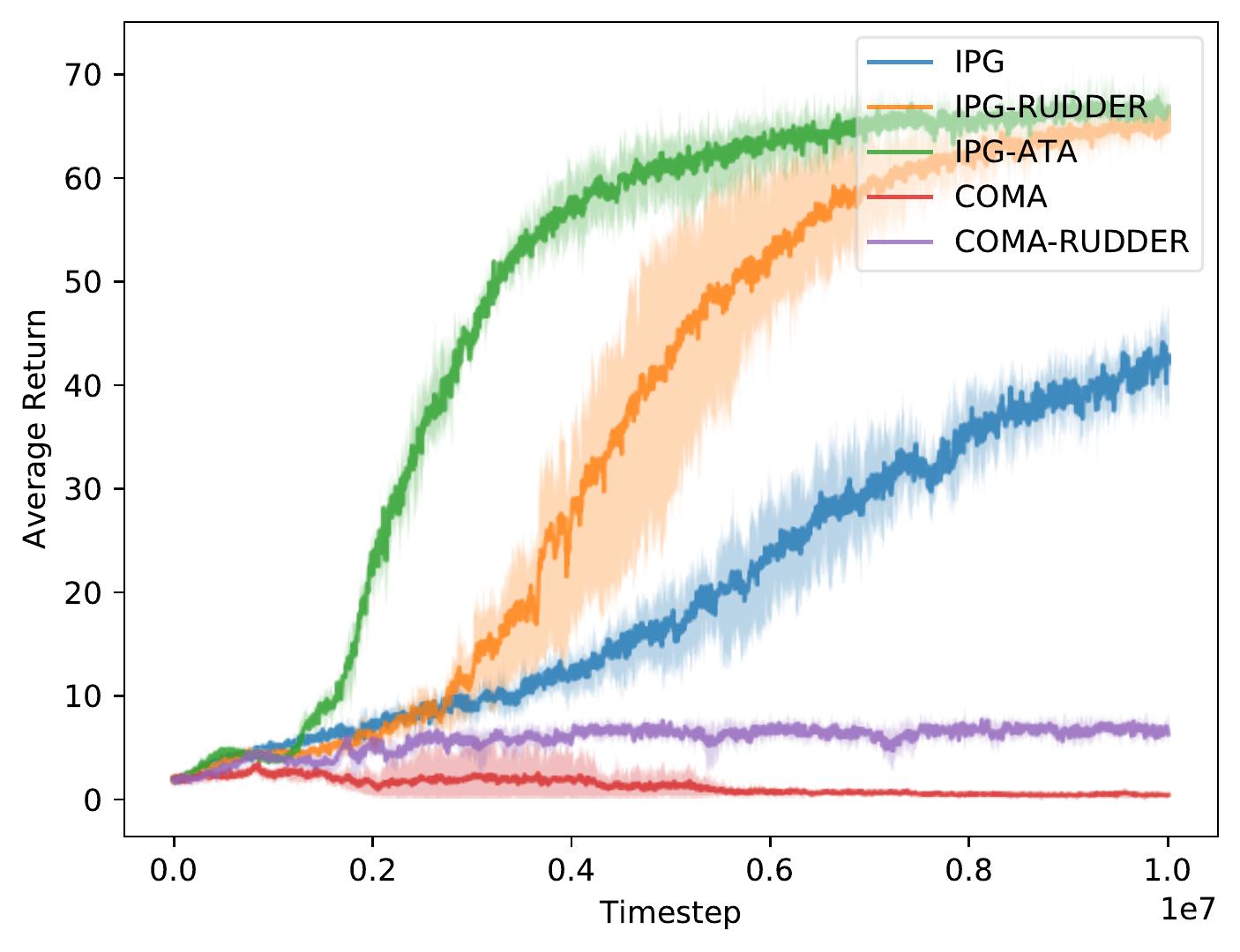}
         \caption{\#Agents = 3, \#Rooms = 2}
     \end{subfigure}
     \begin{subfigure}[b]{0.4\textwidth}
         \centering
         \includegraphics[width=\columnwidth]{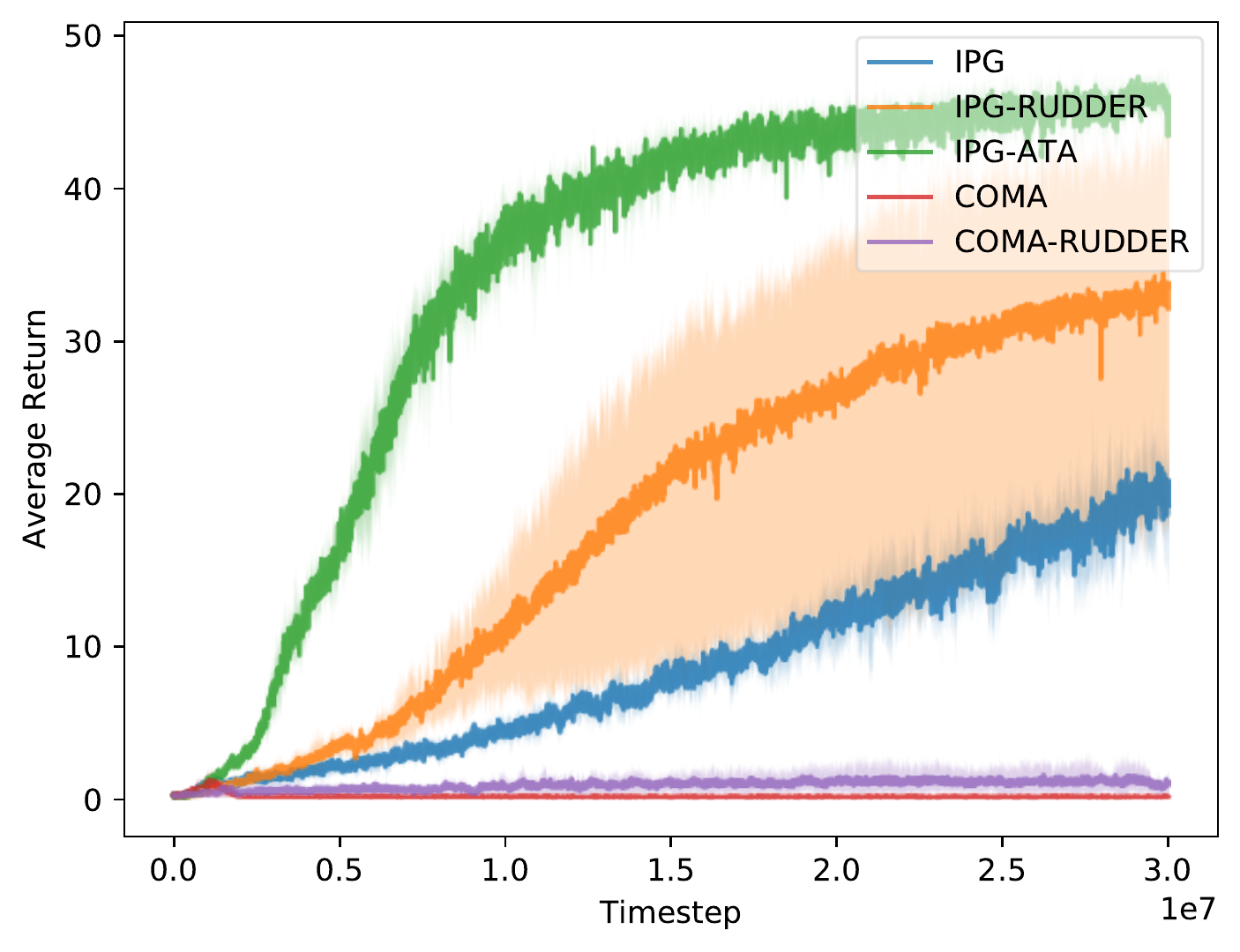}
         \caption{\#Agents = 3, \#Rooms = 3}
     \end{subfigure}
 \caption{2D MultiRoom: average return over training with different number of agents and rooms.}
 \label{fig:2d_mr}
\end{figure*}

\begin{figure*}
     \centering
     \begin{subfigure}[b]{0.4\textwidth}
         \centering
         \includegraphics[width=\columnwidth]{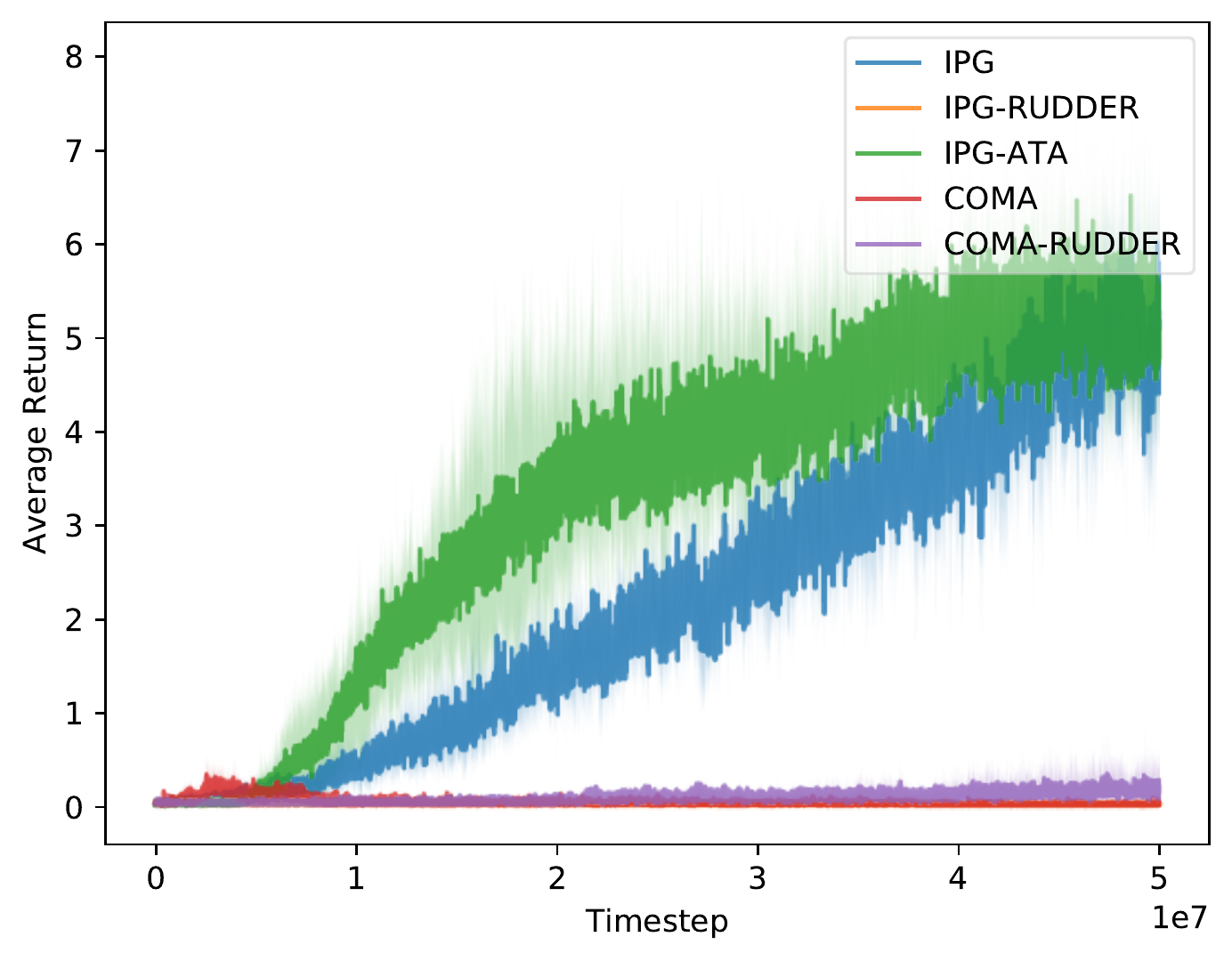}
         \caption{\#Agents = 2}
     \end{subfigure}
     \begin{subfigure}[b]{0.4\textwidth}
         \centering
         \includegraphics[width=\columnwidth]{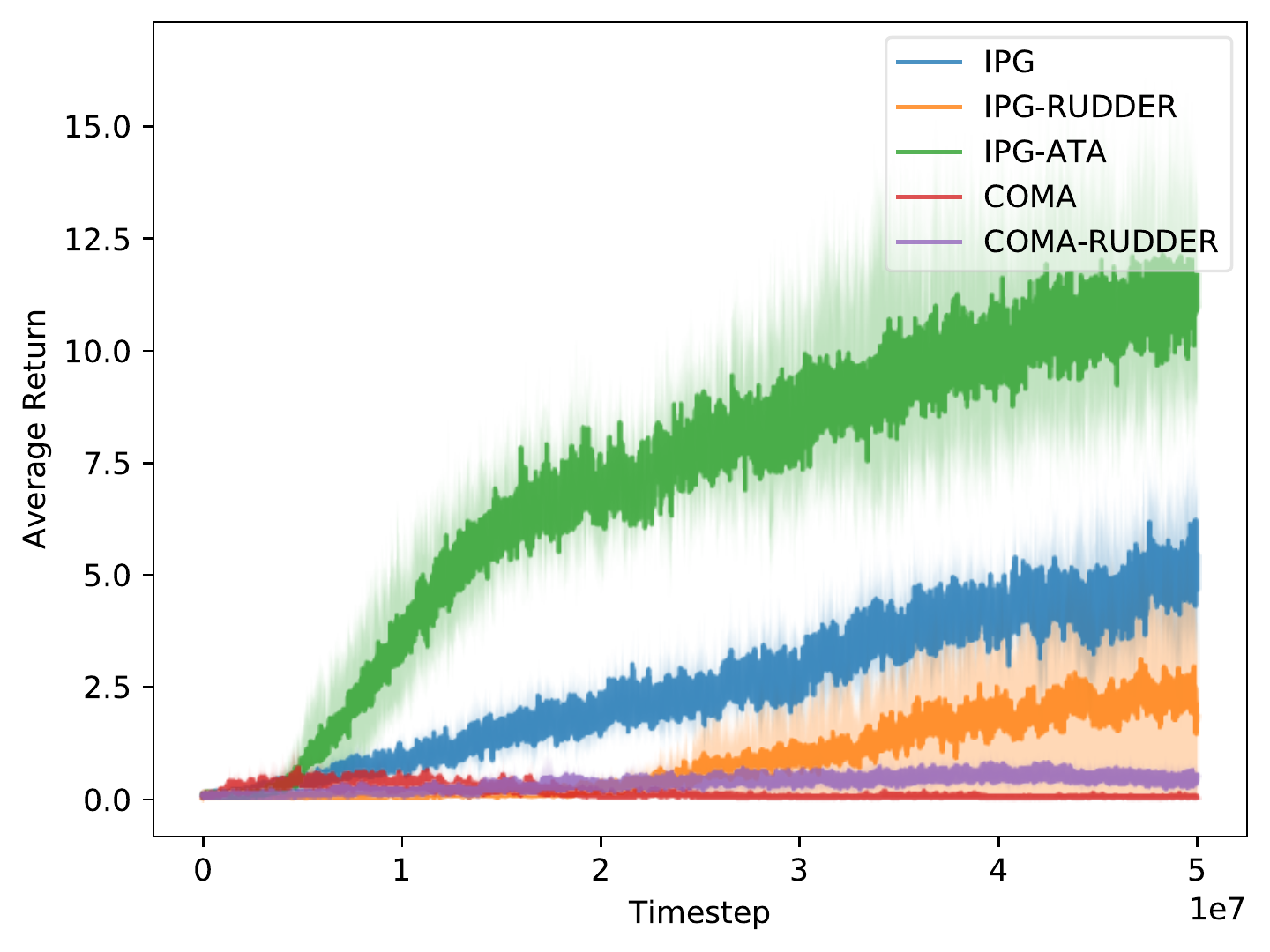}
         \caption{\#Agents = 3}
     \end{subfigure}
 \caption{2D DoorKey: average return over training with different number of agents.}
 \label{fig:2d_dk}
\end{figure*}

Results for MultiRoom and DoorKey-inspired environments are shown in \Cref{fig:2d_mr} and \Cref{fig:2d_dk} respectively. Across most MultiRoom variants and all DoorKey variants, ATA along with PG is clearly the most sample efficient and results in the best policy. This is likely because ATA is able to assign credit well enough through reward redistribution; encouraging agents towards better actions early on in training. 
ATA + PG does not perform well on MultiRoom $m=3$, $n=2$ and we hypothesize that this is due to the lazy versus greedy agent behaviour encouraged by global and local rewards, and can be mitigated by tuning relative magnitudes of global and local prediction losses used to train ATA.

COMA does not perform well in these environments, and we hypothesize that this is due to multiple factors: 1) In sparse rewards settings, the critic in actor-critic based methods learns to predict zero early on in training, and does not provide the actor useful signals to learn as a result. In contrast, the Monte Carlo value estimate provided by vanilla policy gradient is a lot more adaptive to sparse rewards. 2) COMA also gets a boost from the use of value estimates of global state information which is not available in our experiment setting.

\subsection{Ablation Study}
In order to examine how various components of ATA contribute to its performance, we conduct a variety of ablation experiments specifically for MultiRoom $m=3, n=3$. %

\begin{figure}
    \centering
    \includegraphics[width=0.65\columnwidth]{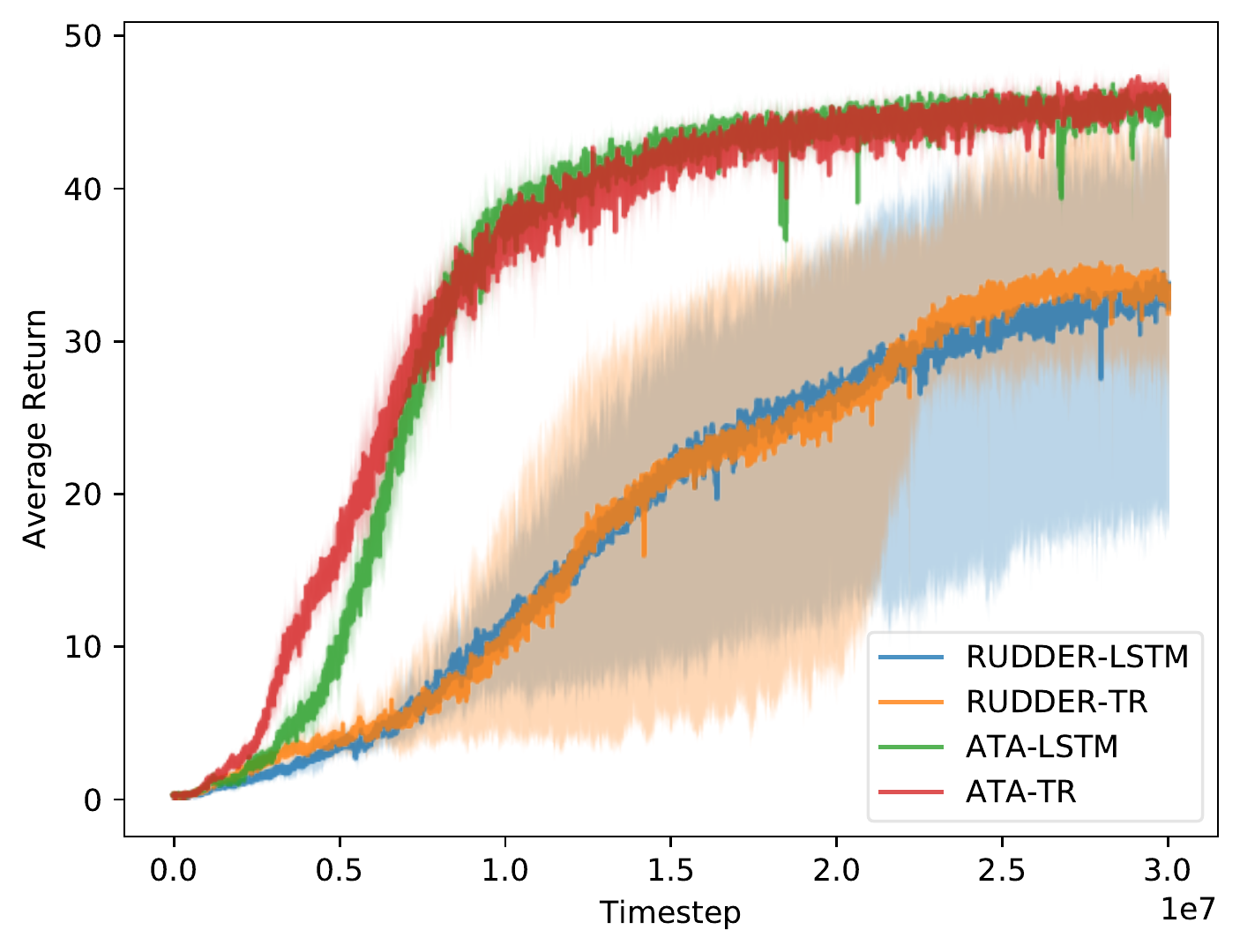}
    \caption{Effect of Architecture -- LSTM vs. Transformer}
    \label{fig:lstm_v_transformer}
\end{figure}
 \begin{figure}
     \centering
     \includegraphics[width=0.65\columnwidth]{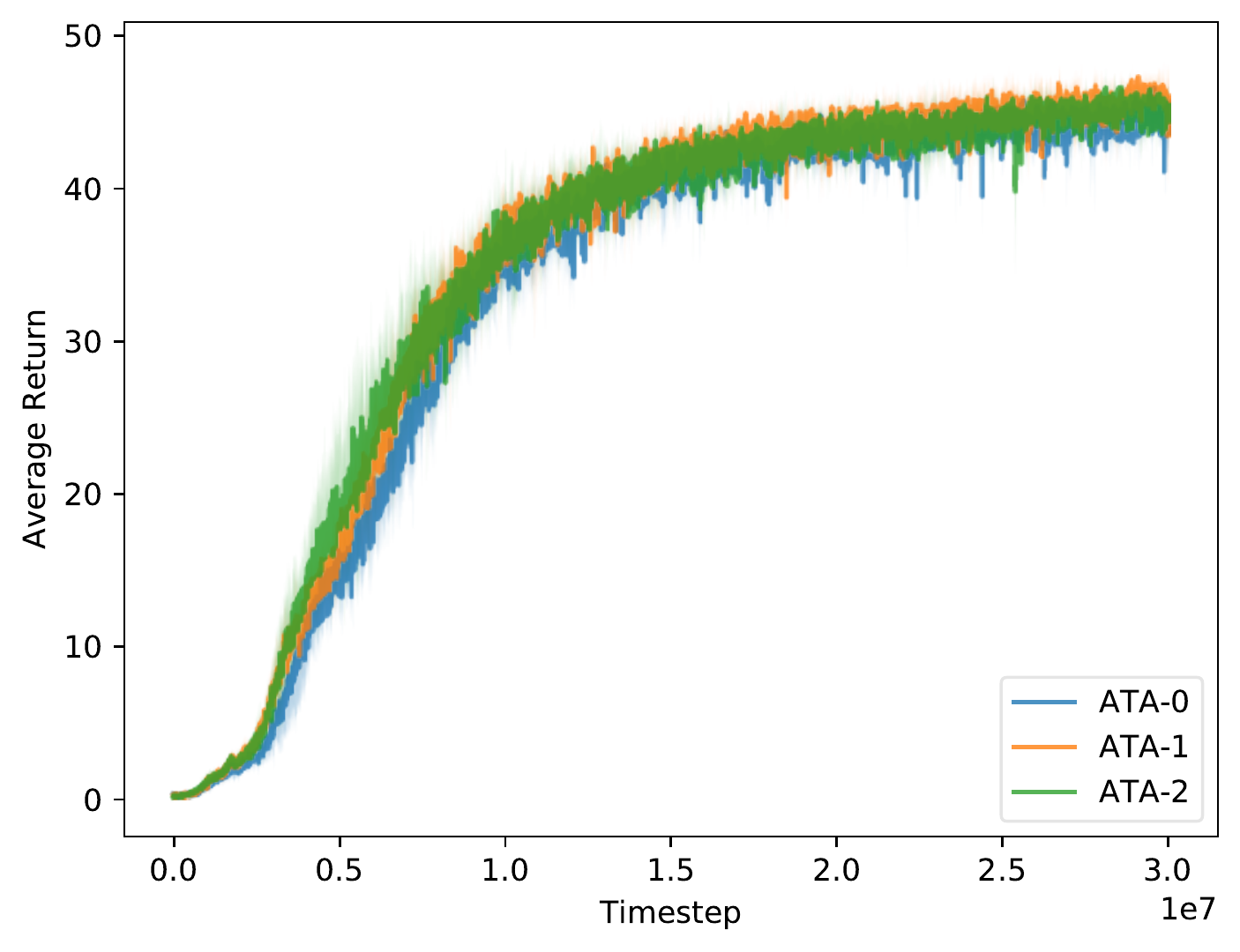}
     \caption{Effect of Architecture -- Mean vs. Attention. 0) mean, 1) linear attention, and 2) multi-layer attention.}
     \label{fig:mean_v_attention}
 \end{figure}
 \begin{figure}
     \centering
     \includegraphics[width=0.65\columnwidth]{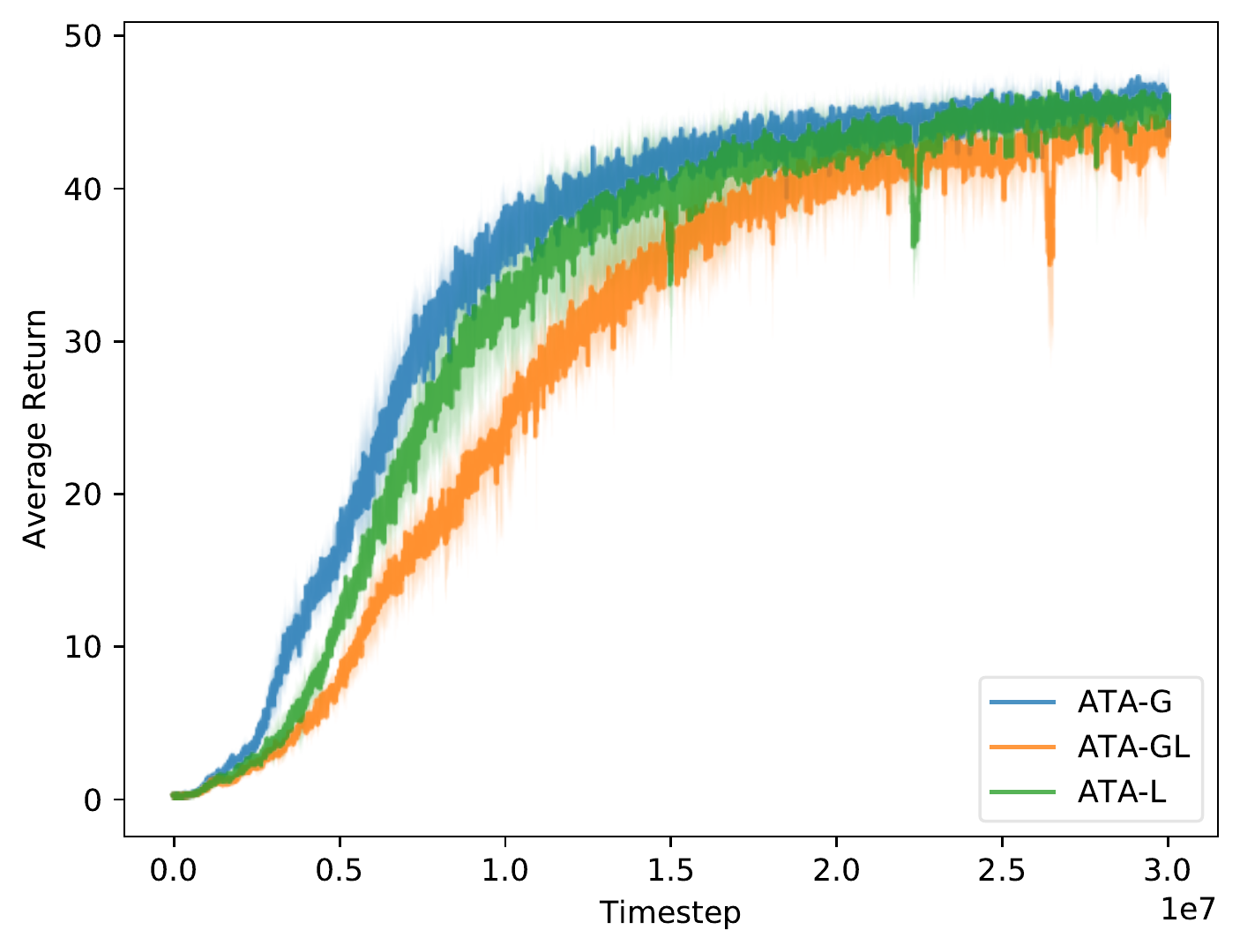}
     \caption{Effect of Prediction Loss -- Global vs. Local: global loss, global + local loss, and local loss respectively.}
     \label{fig:global_v_local_loss}
 \end{figure}
 \begin{figure}
     \centering
     \includegraphics[width=0.65\columnwidth]{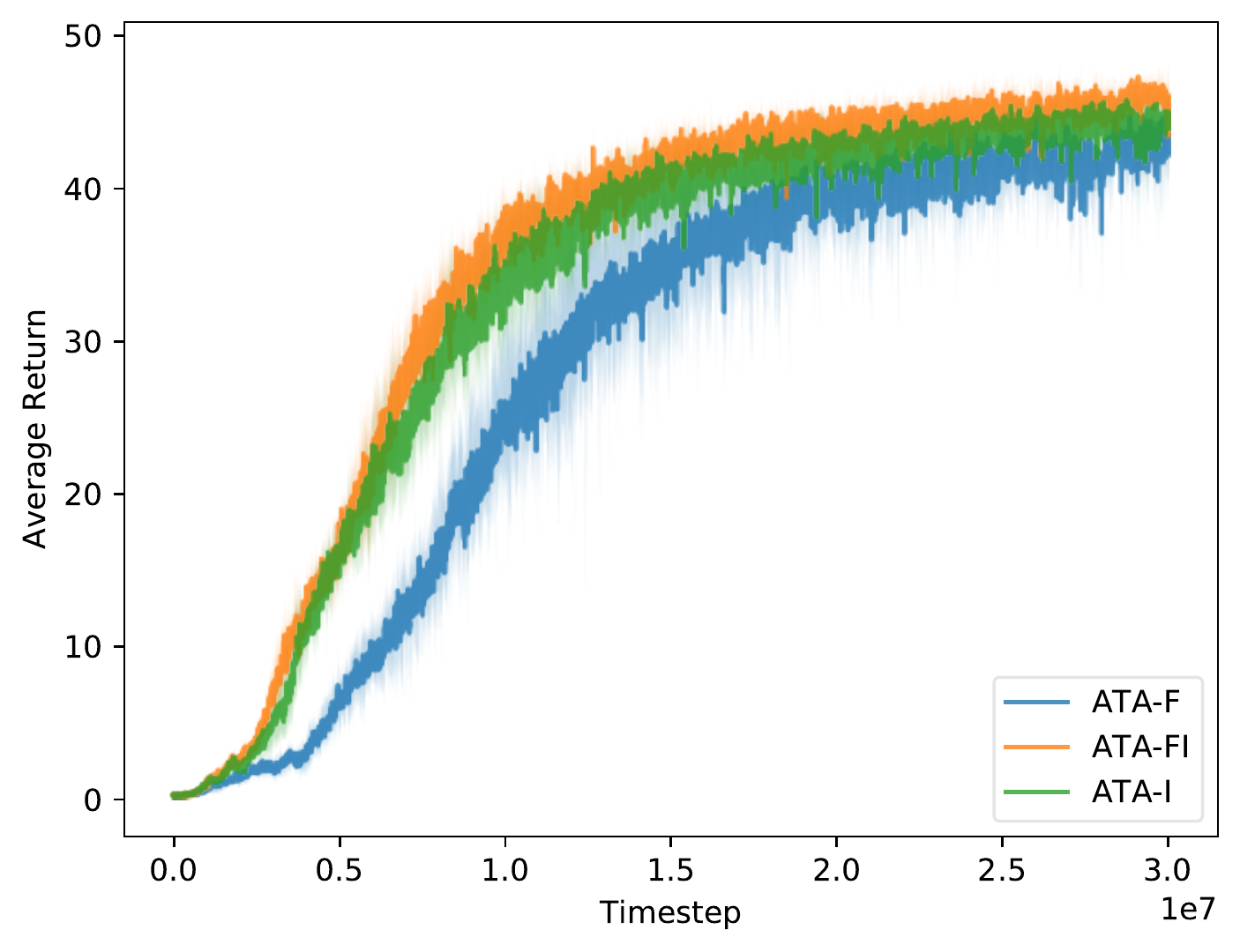}
     \caption{Effect of Timestep Loss -- Intermediate vs. Final: final loss, final + intermediate loss, and intermediate loss respectively.}
     \label{fig:intermediate_v_final_loss}
 \end{figure}

 \begin{figure}
     \centering
     \includegraphics[width=0.65\columnwidth]{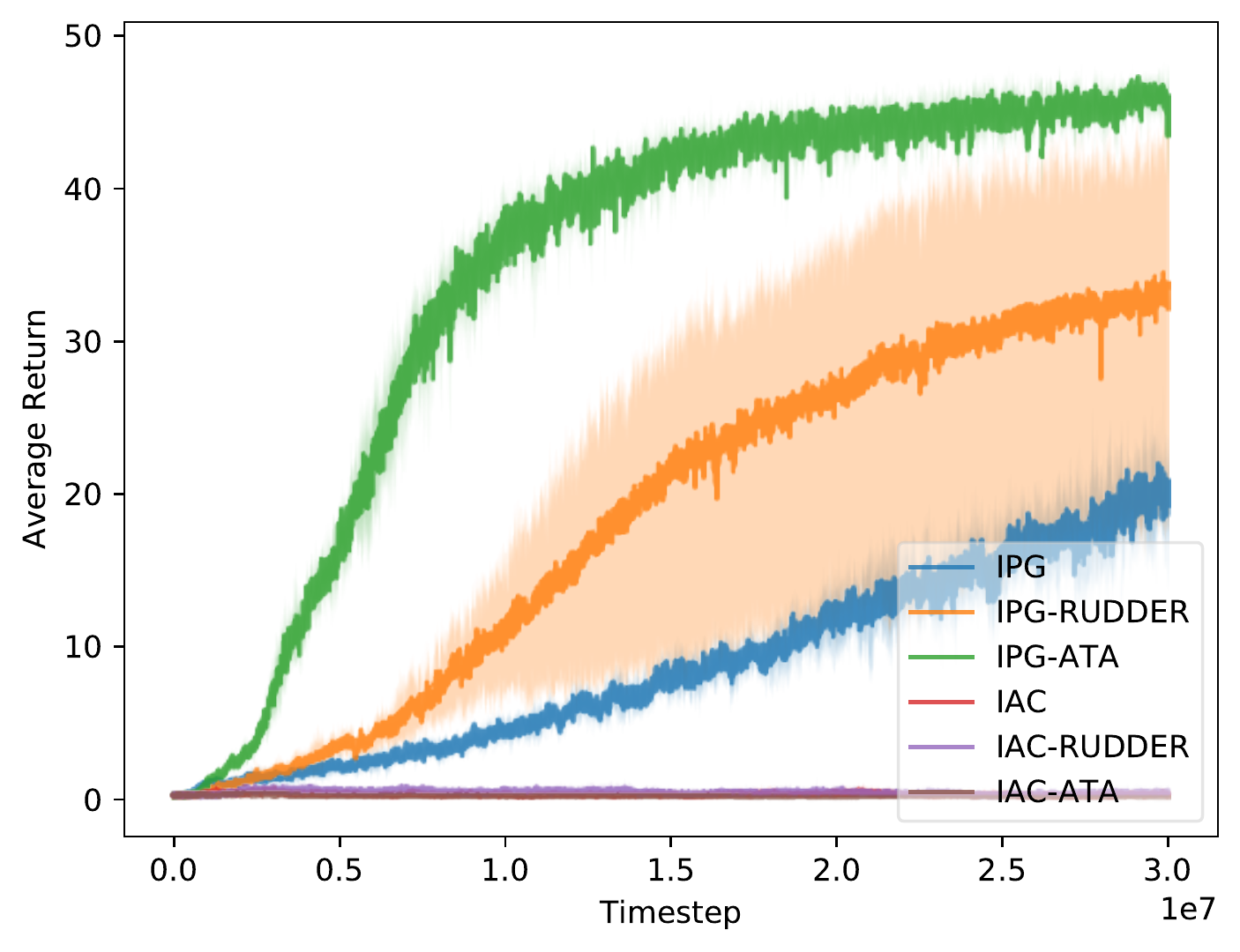}
     \caption{Effect of Value Learning -- IPG vs. IAC}
     \label{fig:ipg_v_iac}
 \end{figure}

\subsubsection{Effect of Architectural Choices} 

\begin{itemize}
\item \textbf{Transformer vs. LSTM Architecture:} We compare the performance of various sequential neural network architectures by replacing the LSTM network inside the RUDDER reward redistribution model with a transformer decoder, and replacing the transformer decoder inside ATA with an LSTM layer. As can be seen in \Cref{fig:lstm_v_transformer}, across both reward redistribution models, the transformer architecture leads to better policy learning early on. In addition, ATA outperforms RUDDER even after accounting for the network architecture. 

\item \textbf{Mean vs. Attention Layers:} We compare the performance of various ways to compute the internal $z^t$ in ATA, from a simple mean to more complicated multi-layer attention. From \Cref{fig:mean_v_attention}, the differences seem minimal and if anything, the simple mean performs the best. We hypothesis that small attention weights in the attention layers may hinder learning the initial agent-specific layers by shrinking gradients, and is likely unnecessary when $n$ is small.
\end{itemize}

\subsubsection{Effect of Different Model Learning Loss variants}

\begin{itemize}
\item \textbf{Global vs. Local Prediction Losses:} We compare the performance of training ATA using various combinations of losses on its global predictions $\hat{R}_t$ and losses on its local predictions $\hat{R}_{it}$ in \Cref{fig:global_v_local_loss}. In this specific setting, defining losses only on $\hat{R}_t$ results in the best performance.
    
\item \textbf{Intermediate vs. Final Timestep Losses:} We compare the performance of training ATA using losses on predictions at all timesteps $t=0, ..., T$ to losses on predictions only at the last timestep $T$ in \Cref{fig:intermediate_v_final_loss}. We see that defining losses on predictions at intermediate timesteps is important in achieving good performance. This is likely because supervising the model to perform well at intermediate timesteps to biases it to redistribute rewards to early timesteps where possible, which results in better credit assignment.
\end{itemize}

\subsubsection{Effect of Value Learning}
\begin{itemize}
\item \textbf{IPG vs. IAC:} We compare the performance of IPG to IAC with and without reward redistribution. IAC performs poorly across all reward redistribution techniques, likely because in this sparse and delayed rewards setting, the critic learns to predict zero early on in training and provides the actor with poor learning signals. This contributes to why COMA also performs poorly across the various environments.
\end{itemize}

\subsection{Limitations}
Sparse reward multi-agent reinforcement learning is a difficult problem with high sample complexity making experiments challenging.
While reward redistribution gives a path to reduce this sample complexity, it comes at an additional computational cost of learning and executing another neural-network model. Depending on the speed of the simulator, these trade-offs may not always be in favor of ATA for wall-clock time of the experiments.
Our ATA training loss adds another hyperparameter to tradeoff between global vs local optimizing behavior.
Full evaluation of its utility requires large scale experiments on a variety of environments with different reward schemes, horizons, observation space and action spaces.

%% file: 05_conclusion.tex
\section{Conclusion \& Future Work}
In this work, we explore the issue of sparse and delayed rewards for multi-agent deep reinforcement learning, which is a relatively less explored area of study.
We proposed Agent-Time Attention (ATA) for redistributing sparse global team rewards into dense agent-specific rewards which can be trained along with existing RL methods such as the standard policy gradient method. 
We extended several single-agent MiniGrid environments to the multi-agent sparse rewards setting, and show that ATA with policy gradient outperforms many strong baselines many of which take multi-agent credit assignment into account on most instances of these environments. 
We also performed multiple ablation analysis experiments to understand the effect of various architecture and training choices in our experiments.
For future work, we would like to extend our model to more complex multi-agent settings, explore more efficient transformer architectures, and end-to-end learn the global versus local rewards trade-off.